REVIEW

# From text to multimodal: a survey of adversarial example generation in question answering systems

Gulsum Yigit[1,2] · Mehmet Fatih Amasyali[1]

**Abstract**
Integrating adversarial machine learning with question answering (QA) systems has emerged as a critical area for understanding the vulnerabilities and robustness of these systems. This article aims to review adversarial example-generation techniques in the QA field, including textual and multimodal contexts. We examine the techniques employed through systematic categorization, providing a structured review. Beginning with an overview of traditional QA models, we traverse the adversarial example generation by exploring rule-based perturbations and advanced generative models. We then extend our research to include multimodal QA systems, analyze them across various methods, and examine generative models, seq2seq architectures, and hybrid methodologies. Our research grows to different defense strategies, adversarial datasets, and evaluation metrics and illustrates the literature on adversarial QA. Finally, the paper considers the future landscape of adversarial question generation, highlighting potential research directions that can advance textual and multimodal QA systems in the context of adversarial challenges.

**Keywords** Question answering · Adversarial question generation · Visual question generation · Adversarial datasets · Adversarial evaluation metrics

## 1 Introduction

In recent years, natural language processing (NLP) has experienced a remarkable transformation through revolutions in deep learning architectures and the availability of vast amounts of textual data. These advancements have led to the development of sophisticated question answering (QA) systems that aim to bridge the gap between human language and machine understanding. State-of-the-art QA models, such as transformer-based architectures like BERT (Bidirectional Encoder Representations from Transformers) and its variants, have

Gulsum Yigit and Mehmet Fatih Amasyali have contributed equally to this work.

✉ Gulsum Yigit
gulsum.yigit@std.yildiz.edu.tr

Mehmet Fatih Amasyali
amasyali@yildiz.edu.tr

[1] Department of Computer Engineering, Yildiz Technical University, Istanbul, Turkey

[2] Department of Computer Engineering, Kadir Has University, Istanbul, Turkey

demonstrated remarkable performance improvements. These models utilize pre-trained language models to capture complex linguistic patterns and contextual dependencies, enabling them to generate accurate and relevant answers to user questions. Moreover, integrating multimodal data, where textual, visual, or audio information joins, has also raised the capabilities of QA systems. Such multimodal QA models, leveraging textual, visual, or audio signals, have shown promising results in analyzing and generating questions. However, a vital vulnerability has arisen in the form of adversarial examples among their outstanding improvements.

Initially presented in computer vision, adversarial examples are meticulously formulated inputs created to fool machine learning models [1]. These inputs are modified in malicious forms, almost invisible to humans but are influential in yielding AI systems to develop inaccurate outcomes. Recent research has brought to light the utilization of adversarial examples in various scenarios. To illustrate, adversaries can simulate physical adversarial examples, thereby confusing autonomous vehicles by manipulating the appearance of a stop sign within a traffic sign recognition system [2, 3]. Additionally, malicious actors show expertise in generating adversarial commands aimed at sabotaging automatic speech recognition systems and voice-controllable systems [4, 5], including crucial platforms such as Apple Siri, Amazon Alexa, etc.

The influence of adversarial examples quickly attracted the researchers' attention, defense mechanisms started to develop, and researchers investigated the underlying causes of the vulnerabilities. With the increased interest in QA systems, the research has shifted to innovative approaches, such as adversarial question generation models, to enhance the quality, diversity, and effectiveness of question generation across textual and multimodal contexts.

Adversarial examples in QA systems are maliciously formulated inputs created to fool the model and generate inaccurate answers. These adversarially designed inputs are constructed by applying subtle transformations to the original question, usually in a form that is invisible to humans but drastically modifies the model's generated answer. These adversarial perturbations could vary from word substitutions to more complicated linguistic structures aimed at making inaccurate or biased texts, images, or videos, challenging the robustness of textual and multimodal QA systems.

The impact of adversarial examples on QA systems performance can be critical. They pose severe challenges to the reliability and security of these systems. Adversarial examples in QA systems can show several damaging effects:

- Decreased Accuracy: Adversarial examples can reduce the accuracy of QA systems. By utilizing vulnerabilities in the model's interpretation of language and reasoning, these malicious inputs can let the model generate inaccurate answers, leading to possible misinformation and degraded user experience.
- Inconsistent Answers: Adversarial examples can also result in inconsistency. A QA system might provide distinct answers for the same question depending on malicious perturbations.

Table 1 shows a comparative overview of existing surveys, focusing primarily on adversarial attacks and defenses across various domains. Wiyatno et al. explored adversarial attacks in the visual domain, investigating different attack methods and defenses [6], while Wang et al. delved into adversarial text generation techniques, presenting a taxonomy of attacks and defenses specific to textual data [7]. Xu et al. summarized studies on adversarial examples across different data types, discussing threats and countermeasures [8]. Huq et al. systematically analyze various adversarial attacks and defense strategies on text, including classification, machine translation, and question answering. It synthesizes findings to identify

**Table 1** Comparison of existing surveys

| Year | Existing survey | Broad topics |
|---|---|---|
| 2019 | [6] | Focuses on adversarial attacks in the visual domain, exploring various attack methods, defenses, and their effectiveness to provide extensive field coverage |
| 2019 | [7] | Classifies adversarial techniques for crafting adversarial texts, presents taxonomy of attacks and defenses in the text domain, discusses challenges, and future research directions |
| 2020 | [8] | Summarizes studies on adversarial examples and their countermeasures across different data types, discussing threats, countermeasures, and future directions |
| 2020 | [9] | Systematically analyzes various adversarial attacks and defense strategies on texts, including classification, machine translation, and question answering, and highlights challenges for future research |
| 2021 | [10] | Provides a taxonomy of question generation tasks, analyzes existing models, compares them through benchmark tasks, and discusses future directions |
| 2021 | [11] | Unifies definitions, evaluations, and mitigation strategies for robustness in NLP, connecting various lines of work and outlining open challenges and future directions |
| 2023 | [12] | Review of techniques for question generation, evaluation metrics |
| 2023 | [13] | Reviews studies on adversarial attacks in medical imaging, highlighting concerns and potential risks in deep learning technology in medicine |

**Table 1** continued

| Year | Existing survey | Broad topics |
|---|---|---|
| 2023 | [14] | Offers a thorough examination of adversarial defense strategies in NLP across various tasks, also introduces a new classification system for these defense methods, and addresses ongoing issues in this field |
| 2024 | Our Review | Comprehensive analysis of adversarial example generation techniques in QA, including textual and multimodal contexts (2017–2024). Reviews defense mechanisms, adversarial datasets, evaluation metrics analysis, and future research directions |

advantageous techniques and highlight challenges for future research [9], and Zhang et al. provided a taxonomy of question generation tasks, analyzing existing models and comparing them through benchmark tasks [10]. Mulla et al. reviewed techniques for question generation and evaluation metrics [12], while Sorin et al. reviewed adversarial attacks in medical imaging [13]. Goyal et al. offer an examination of adversarial defense methods across various text-based tasks introducing a classification system for these defense methods and address ongoing issues in this field [14].

QA systems are widely used in various fields, including healthcare, finance, customer service, education, etc. However, they are vulnerable to adversarial attacks, risking misdiagnoses, financial losses, etc. Ensuring their robustness against adversarial attacks is crucial to preserve the trustworthiness of these systems. While there may be existing research on adversarial attacks in the broader context of NLP or image-based tasks, a focused survey on QA systems is essential to examine the current state of research in this specific domain. Moreover, adversarial attacks and defenses in QA systems are rapidly evolving, with new techniques and methodologies emerging constantly. Therefore, there is a need for an updated assessment of the current state of the art. In this study, we aim to fulfill this need. While we thoroughly examine the approaches used, we acknowledge the significance of placing them in the context of existing research.

In this survey, we aim to thoroughly investigate and review adversarial example generation, specifically in QA systems, covering both textual and multimodal contexts from 2017 to 2024, and categorize the diverse techniques utilized in this field. By classifying the various approaches, we provide a structured interpretation of how adversarial examples manipulate these systems. Furthermore, we examine defense mechanisms that enhance robustness against adversarial attacks, review datasets, and evaluation metrics. Lastly, we discuss potential future research directions in adversarial question generation for textual and multimodal QA systems. Consequently, our paper aims to contribute to the continuing research about adversarial inputs in QA systems, offering insights into current research directions and suggesting further investigation to improve the security and reliability of QA systems.

The paper is organized as follows. Section 2 first provides an overview of traditional QA systems. Subsequently, we explore the existing approaches for generating adversarial examples in QA systems, examining rule-based perturbations and the more sophisticated generative models that have gained prominence in recent research. Section 3 analyzes the landscape of multimodal QA systems, beginning with an overview of the complex interactions between different modalities. We then navigate the existing multimodal question generation approaches, including generative models, encoder–decoder architectures (Seq2Seq), and hybrid methodologies. Section 4 delves into the diverse defense mechanisms designed for QA systems and reviews the adversarial datasets and evaluation metrics that quantify the impact of adversarial perturbations on QA systems performance, addressing both textual and multimodal aspects. Finally, in Sect. 5, we discuss future research direction in adversarial QA.

**Research Strategy:**

Figure 1 provides a visual overview of the investigation and elimination criteria used in the review process. The search was conducted across multiple databases, including ACM Digital Library, IEEE Xplore, Springer Link, and Scopus, from 2017 to January 2024. The search terms employed were "adversarial," "example," "generation," "attack," and "question answering." The selection of articles for this study followed a two-level inclusion and exclusion process.

Initially, 557 articles were identified, with 527 sourced from database searches and 30 identified through other resources. After removing duplicate entries, 178 unique studies remained for further evaluation. The screening process excluded 69 studies that did not align with the review's primary focus, which was the creation of adversarial examples specifically targeting QA systems. The remaining 109 studies underwent a thorough assessment for eligibility. Among these, 76 were excluded with reasons. The primary reasons for exclusion included a lack of relevance to adversarial example generation for QA systems, limited availability of full-text articles for thorough evaluation, non-English language publications, and studies focusing on different target systems or tasks, such as image classification, image captioning, or NLP tasks unrelated to QA systems. Accordingly, 33 studies were included in the systematic review.

## 2 Adversarial example generation in question answering systems

### 2.1 Overview of question answering systems

QA is a complex and challenging problem within NLP, encompassing tasks like comprehending human language, generating responses, and representing information about the world. The primary aim is to automatically provide accurate answers to questions in natural language by extracting the most relevant information [15–18]. Various QA systems have attempted to respond to user queries through various approaches. These implementations have tackled an exhaustive exhibition of domains, databases, question formats, and answer structures [19].

Deep learning architectures such as convolutional neural networks (CNNs) and recurrent neural networks (RNNs) are broadly used in QA systems to capture complicated patterns within textual data and comprehend complex relationships between words and phrases [20, 21]. Seq2seq models, a famous architecture in NLP, enable the transformation of input sequences (questions) into output sequences (answers). These models have significantly improved in generating coherent, contextually appropriate responses [22–24].

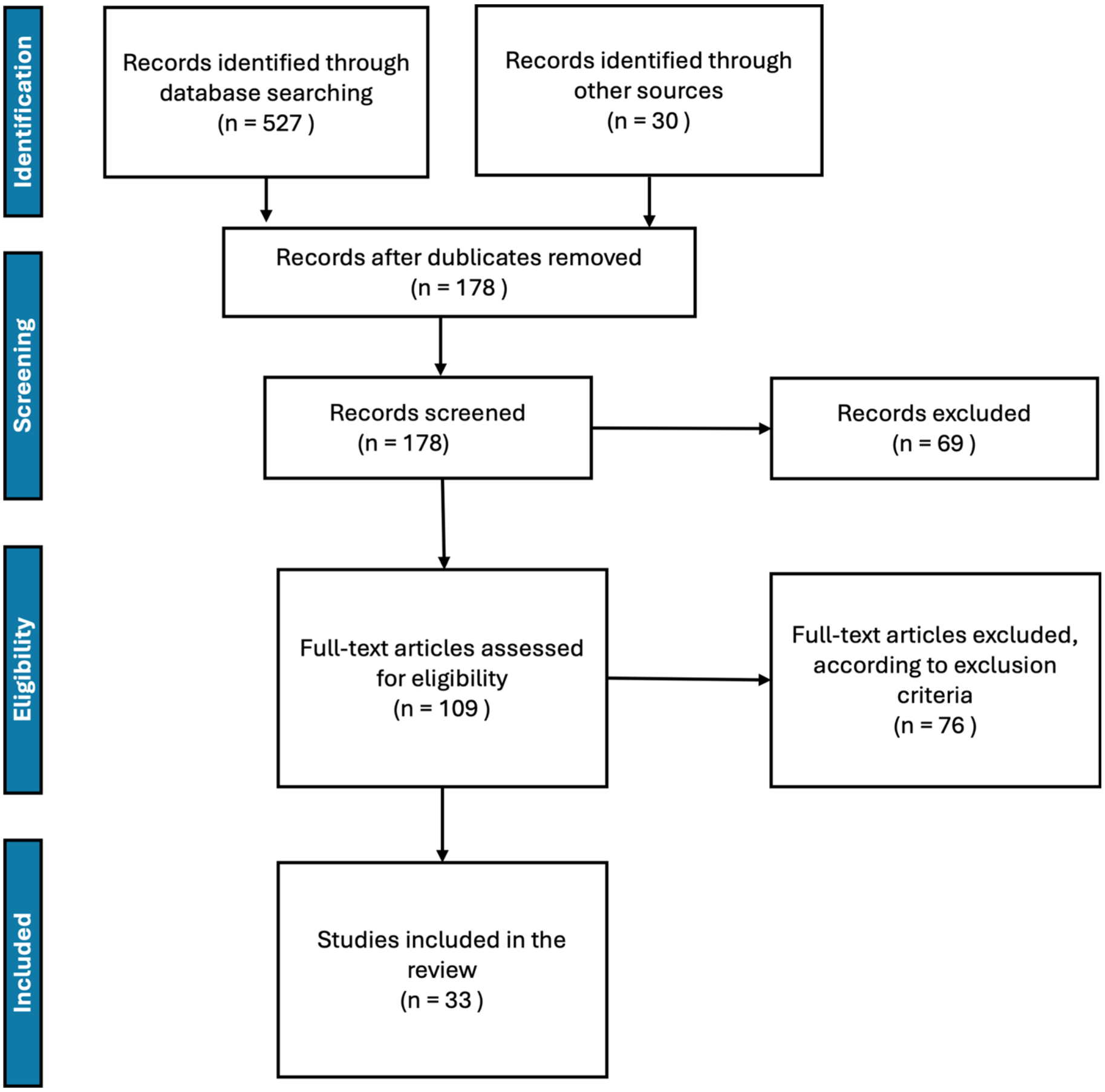

**Fig. 1** Research strategy

Recent QA systems capitalize on advanced techniques that extend beyond those mentioned. Attention mechanisms, a core component in modern neural networks, enable systems to focus on specific parts of input data, enhancing their ability to determine crucial information in questions and contexts. Attention mechanisms allow QA systems to weigh the importance of different words in a question or context, leading to more accurate and contextually relevant answers [25–27].

Ensemble learning further contributes to QA system robustness by combining predictions from multiple models. This technique enhances the performance and reduces the risk of relying on a single model's biases or errors [28, 29]. On the other hand, transfer learning involves training models on large datasets and then fine-tuning them on specific QA tasks, enabling them to learn from wide linguistic contexts and specialize in domain-specific questions [30–32]. Furthermore, reinforcement learning techniques have been explored to enhance QA systems. Through trial and error, these models learn to optimize their responses by receiving rewards for generating accurate answers and penalties for incorrect ones. This approach enhances the learning process, enabling systems to iteratively improve their performance based on user feedback or predefined evaluation metrics [33–35].

Recently, Large Language Models (LLMs) have gained significant interest. These models, pre-trained on vast textual corpora, maintain an understanding of linguistic structures and context. They are fine-tuned or prompted for QA tasks to effectively analyze and interpret questions before generating relevant responses [36–38]. From attention mechanisms and seq2seq models to ensemble learning, transfer learning, LLMs, and reinforcement learning, these approaches collectively contribute to the performance of QA systems in comprehending and responding to user questions across diverse contexts.

### 2.2 Existing approaches of adversarial example generation in question answering systems

This section examines various strategies used to craft adversarial examples within textual QA systems. We categorize existing approaches to generating adversarial examples in textual QA systems, and our classification criteria are based on the methodologies employed to craft adversarial examples. Specifically, we classify the approaches into rule-based perturbations and generative models. Rule-based perturbations concern systematically manipulating questions' structure, grammar, or wording using predefined rules to generate adversarial examples. On the other hand, generative models leverage methods such as deep learning architectures, seq2seq models, Generative Adversarial Networks (GANs), and autoencoders to generate contextually suitable adversarial questions. By categorizing the approaches, we aim to give a precise understanding of the various methods employed in generating adversarial examples within QA systems.

#### 2.2.1 Rule-based perturbations

Rule-based perturbations concern systematically manipulating questions' structure, grammar, or wording using predefined rules to generate adversarial examples. By applying syntactic or semantic modifications to the original question texts, tricky variations can be constructed, which leads to probable vulnerabilities in QA systems. These perturbations are vital in assessing the robustness and reliability of QA systems. Figure 2 visually demonstrates how rule-based perturbations are applied to the original question to create an adversarial input. This technique applies predefined rules or modifications to the original question. These perturbations might not be immediately apparent to humans but can mislead or confuse the model. Once the adversarial input is crafted, it is given to the model. The model processes this adversarial question just like any other input. However, the generated output is incorrect due to the strategically introduced perturbations that exploit the model's weaknesses.

In [39], two methods for generating adversarial examples for Math Word Problem (MWP) solvers are introduced. The first technique, "Question Reordering," considers reordering the question part of the problem to the beginning of the problem text. The second technique, "Sentence Paraphrasing," concentrates on rephrasing each sentence in the problem while preserving both the semantic meaning and numeric data. In [40], Wang et al. proposed a new adversary example generation algorithm called "AddSentDiverse" that significantly increases the variance within adversarial training data. The "AddSentDiverse" algorithm is a modified version of "AddSent" [41]. The "AddSent" algorithm involves semantic modifications to the question using antonyms and named-entity swapping. It then generates a fake answer matching the style of the actual answer and integrates it with the modified question to create a distractor. Besides, the "AddSentDiverse" algorithm includes two modifications: randomizing distractor order and diversifying the set of fake answers.

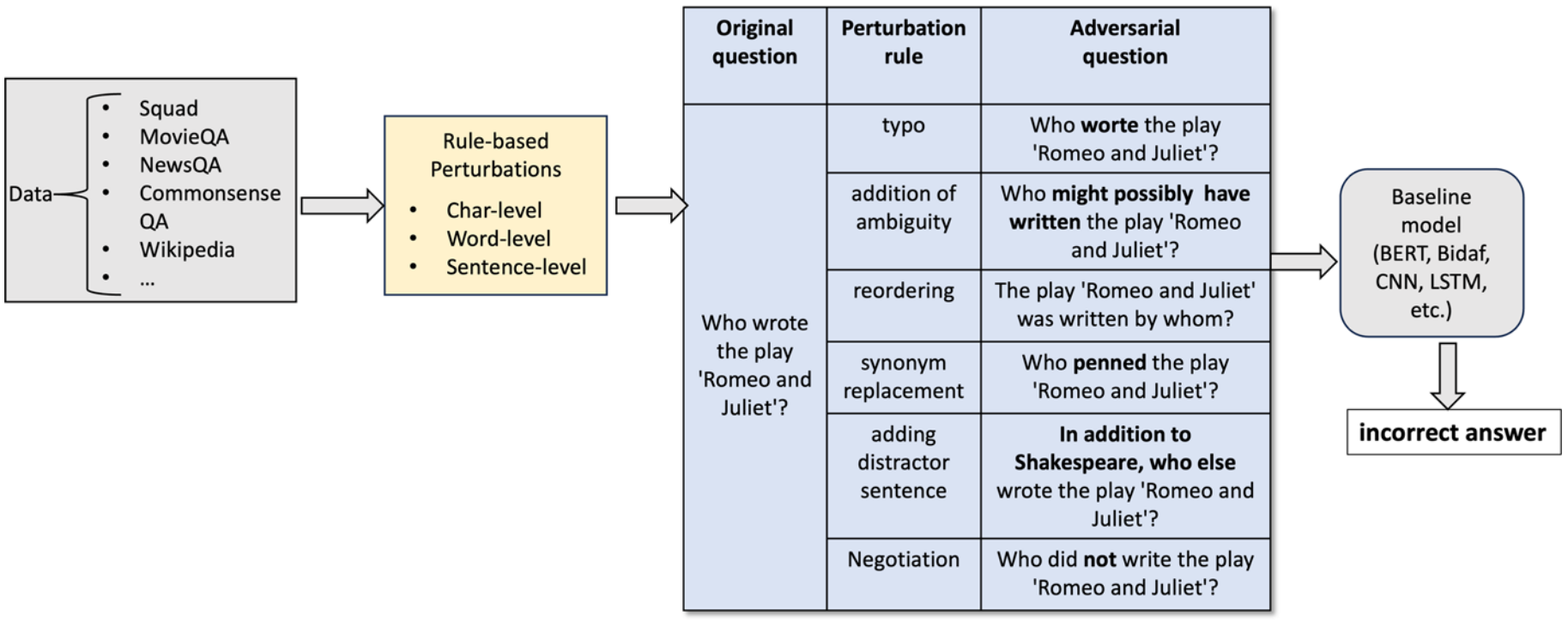

**Fig. 2** Rule-based perturbations on textual data

The study in [42] involves training models on artificially generated adversarial examples. The misleading texts generated by these adversarial examples are created using different methods, including AddSent and AddAny. AddAnyExtend, which expands AddAny, involves an extended vocabulary containing question words, high-frequency words, passage words, and random common words. AddAnsCtx generates misleading text by removing answer words from answer sentences.

Rosenthal et al. [43] extended the AddSentDiverse algorithm [40], which creates adversarial distractors by transforming a question into a statement using predefined rules. The attack proposed is semantically similar to the question but aims to make a wrong statement for humans. Constructed sentences desire to be grammatically correct but are not strictly enforced. The authors developed four attack strategies that generate adversarial statements to confuse the QA system. Additionally, they investigated generating adversarial statements by translating them into six other languages in the dataset. The authors noted that it is suitable for an assessment of the robustness of the model in different languages.

Cao et al. proposed Twin Answer Sentences Attack (TASA), an automatic black-box adversarial attack that changes the context without affecting its fluency or correct answer [44]. TASA recognizes the relevant answer sentence in the context and creates two modified sentences to exploit biases in QA models. One sentence preserves the meaning of the correct answer but substitutes shared keywords with synonyms. Thus, it forms a perturbed answer sentence to reduce the model's focus on the correct answer. The other sentence maintains the keywords and syntax but changes associated entities, creating a distracting answer sentence to mislead the model toward an incorrect answer with irrelevant entities.

Xue et al. introduced an approach called Dependency Parse-based Adversarial Examples Generation (DPAEG) [45]. DPAEG first uses a new dependency parsing-based algorithm to extract keywords from a given question. Then, based on the extracted keywords, the algorithm generates adversarial words, including misspellings and words similar to keywords. These adversarial words are used to form a series of adversarial questions. These new questions are similar to the original questions and do not affect human understanding.

The authors in [46] used adversarial examples to determine deficiencies in language models and understand how they work. They kept the question the same but modified the context information. Various attacks have been implemented, such as sentence reordering, where the sentences of the paragraph are reordered to preserve the semantic meaning of the paragraph, and splitting key sentence attacks, where the critical sentence that is important to the answer is split into two or more sentences, garbage concatenation where "xxxxx" (garbage value) added to the critical sentence of the paragraph.

The study in [47] targeted neural reasoning architectures. The study determines metamorphic relations, which are logical relationships between a system's inputs and outputs. In the QA systems, these relations are established between original questions and their corresponding answers in a neural reasoning system. The perturbations are crafted in a manner that aims to use the vulnerabilities of the system's architecture, yielding it to produce incorrect or inconsistent answers. The perturbed questions and their original answers form the adversarial examples.

The study in [48] introduced an approach for assessing statistical biases in Machine Reading Comprehension (MRC) models. This was done by expanding the answer options for each question with irrelevant options. These unrelated options were randomly selected from the RACE dataset. These questions were unrelated to the current passage and varied from the original answer choices. Each option was then independently scored using a specific procedure. The model prompts to assign likelihood scores to these irrelevant options as potential correct answers. If the model is biased, it might consistently give higher scores to specific irrelevant options, even when all options are irrelevant. These frequently chosen irrelevant options are named "magnet options."

Sun et al. examined the vulnerability of BERT (Bidirectional Encoder Representations from Transformers) to misspellings and presented a strategy called "Adv-BERT" [49]. These perturbations are designed to yield BERT to generate incorrect or inconsistent results. The "Input Reduction" technique is introduced by Feng et al. [50], which considers a step-by-step process of eliminating less significant words from a question. In other words, the method removes unimportant words according to their saliency scores while ensuring the model's output remains unchanged. Removing tokens from the question displays a less specific, unanswerable question, yet the model keeps its original answer due to a spurious correlation.

### 2.2.2 Generative models

Generative models have become useful tools for adversarial question generation within QA systems. This approach leverages various techniques, including deep learning architectures, seq2seq models, GANs, and autoencoders. Seq2seq models help the generation of contextually relevant texts by mapping input sequences to output sequences. Thus, it makes them suitable for crafting adversarial questions. Deep learning architectures generate questions using complex neural networks that capture complicated patterns and semantics. GANs contain two parts: a generator creates questions, and a discriminator evaluates question quality over time. Autoencoders help create adversarial questions by encoding input questions and decoding them with subtle perturbations. Collectively, these generative models contribute to developing more robust QA systems by gradually generating adversarial texts.

Zhu et al. presented SAGE (Semantically valid Adversarial GEnerator) is a white-box attack model, which means it has access to the gradients of the target systems to generate adversarial questions at the sequence level [51]. The model has been built upon the stochastic Wasserstein seq2seq model to generate fluent questions. The work at [52] introduced a framework called T3 to create controlled adversarial attacks in sentiment analysis and QA tasks. The framework uses a tree-based autoencoder to transform the discrete text into a continuous representation. In addition, a new tree-based decoder is used to ensure that the generated text is syntactically correct and allows manipulation at both the sentence (T3(SENT)) and word (T3(WORD)) levels.

In the study by Gan et al. two different test sets are created, consisting of paraphrased SQuAD questions [53]. The first set contains paraphrased questions similar to the original ones. The second set consists of questions that aim to mislead the models and are paraphrased using contextual words and wrong-answer candidates. The authors followed an approach that involves a transformer-based paraphrase model that creates multiple paraphrased versions of a source question using a series of paraphrasing suggestions. The study in [54] applies an adaptable adversarial framework named DQAA, using an efficient GAN-inspired cue finder to identify vulnerable words. DQAAT consists of three essential components. Firstly, a cue finder is responsible for choosing tokens to serve as vulnerable cues. Secondly, a text transformer adjusts sentences using cues that were identified earlier. Lastly, a grammar fixer employs a masked language model to correct possible grammatical errors.

In [55], Blohm et al. prepared two white box attacks to compare the robustness of CNN and RNN. They used the model's internal attention to identify a key sentence to which the model gave significant weight in choosing the correct answer. In the attack, they replaced the most striking words with randomly selected words from a known vocabulary.

## 2.3 Discussion

Investigating adversarial example generation in QA systems reveals various strategies for challenging the performance of these systems. From rule-based perturbations to generative models, researchers use different techniques to create adversarial examples that use vulnerabilities in QA systems' understanding and response mechanisms.

Table 2 examines various studies of adversarial question generation systems. The table delves into critical aspects such as the chosen dataset, the applied approach, original examples, adversarially generated examples, and the extent of multilingual support.

Table 3 provides an illustration and comparison of various key aspects of the study. The table contains details related to the technique, the baseline model used, the dataset utilized for experimentation, the chosen evaluation metric to assess performance, and the baseline and adversarial performance achieved. The table clearly demonstrates that adversarial examples notably impact the performance of baseline models across diverse datasets and evaluation metrics.

Rule-based perturbations aim to create tricky question variations using vulnerabilities in QA systems. Based on the findings of the studies provided, we can deduce the following outcomes.

- Studies have shown that even minor changes to the original questions, such as question reordering or sentence paraphrasing, can significantly impact the performance of QA systems, emphasizing their vulnerability to crafted inputs.
- Techniques like AddSent and AddAny have been suggested to generate adversarial examples by introducing semantic modifications or removing answer words from sentences.
- Extensions of these techniques, such as AddSentDiverse and TASA, aim to improve the variance within adversarial training data and exploit biases in QA models.
- Studies such as those by Kumar et al., Wang et al., and Rosenthal et al. introduce new algorithms to create adversarial examples tailored to specific QA tasks, from MWPs to multilingual QA systems [39, 40, 43]. These techniques exploit weaknesses in QA models' understanding of context, semantics, and syntactic structures.
- Adversarial attacks based on rule-based perturbations can be multilingual and may generate adversarial information by translating questions into other languages.

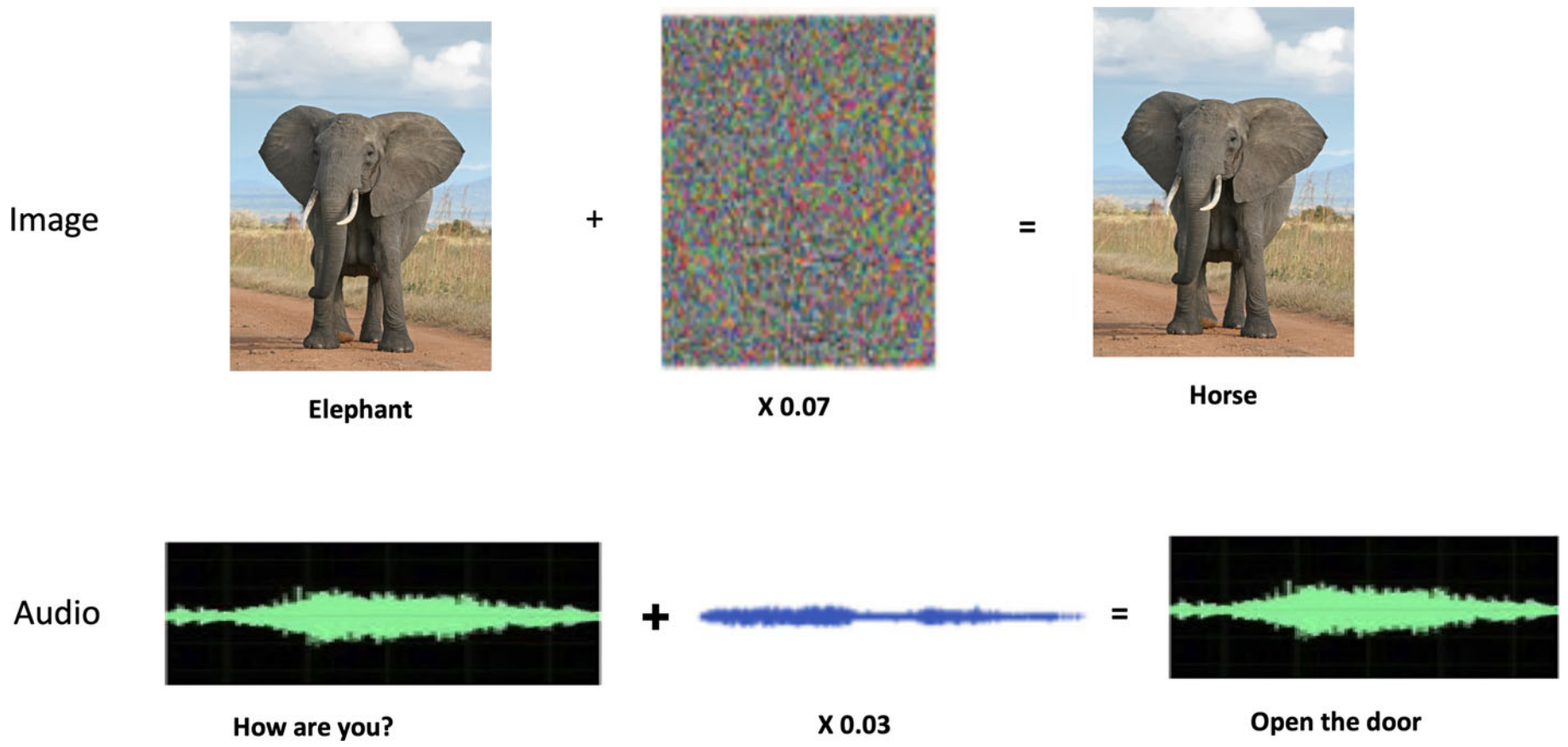

**Fig. 3** Multimodal adversarial example generation: perturbed image and distorted audio

Generative models leverage complex neural networks to generate contextually relevant adversarial questions. Based on the findings of the studies reviewed, we can infer the following outcomes.

- Adversarial attacks based on generative models may involve paraphrasing questions to mislead QA models or identifying vulnerable words using GAN-inspired cue finders.
- Techniques like SAGE and T3 have been introduced to create controlled adversarial attacks by transforming discrete text into continuous representations and ensuring syntactical correctness.
- Studies have compared the robustness of different QA models against adversarial attacks generated by generative models, revealing significant performance degradation in many advanced QA models.

These synthesized insights highlight the effectiveness of both rule-based perturbations and generative models and contribute to developing more robust QA systems by gradually generating adversarial texts and exposing weaknesses in existing models.

# 3 Multimodal question answering systems

## 3.1 Overview of multimodal question answering systems

Multimodal QA systems offer a procedure that includes multiple modalities, such as text, images, videos, and audio. Each modality has its own challenges and requires specific techniques for effective integration. NLP techniques are used to generate questions to understand text-based modalities. Computer vision techniques process visual information and generate appropriate questions for image-based modalities. Other modalities, such as video and audio, pose more challenges due to their temporal nature. They may require procedures such as video summarization and audio transcription to compose the questions. Integrating different modalities introduces various challenges as follows:

- Alignment of data in different modalities: Text data are inherently sequential, while image, video, and audio data naturally have spatial and temporal properties.

**Table 2** Comparative analysis of adversarial question generation systems—dataset, approach, original and adversarial examples, and multilingual support

| Study | Dataset | Approach | Original example | Adversarial example | Multilingual support |
|---|---|---|---|---|---|
| [39] | MaWPS ASDIV-A | Question Reordering | Question:Tim has 5 books. Mike has 7 books. How many books do they have together?<br>Equation: $X = 5+7$ | Question: How many books do they have together given that Tim has 5 books and Mike has 7 books<br>Equation: $X = 5*7$ | no |
| [39] | MaWPS ASDIV-A | Sentence Paraphrasing | Question: "Tim has 5 books. Mike has 7 books. How many books do they have together?<br>Equation: $X = 5+7$ | Question: Tim has got 5 books. There are 7 books in Mike's possession. How many books do they have?<br>Equation: $X = 5*5$ | no |
| [44] | SQuAD 1.1 NewsQA NaturalQuestions HotpotQA TriviaQA | Twin Answer Sentences | Context: …On 17 May 1899, Tesla moved to Colorado Spring …very thick and noisy. He investigated atmospheric electricity, observing lightning signals via his receivers. He stated that … | Context: …very thick and noisy. Tesla looked into atmospheric electrical energy, observing lightning signals via his receivers. He stated that …Charlie investigated static electricity, observing noticeable phenomenon via his receivers | no |
| [45] | WebQuestionsSP CuratedTREC WikiMovies | Typo | Question: Who painted Olympia? | Question: Who printed Olympia? | no |

**Table 2** continued

| Study | Dataset | Approach | Original example | Adversarial example | Multilingual support |
|---|---|---|---|---|---|
| [53] | SQuAD | Paraphrasing | Context: 826 Doctor Who instalments have been televised since 1963 …Starting with the 2009 special "Planet of the Dead," the series was filmed in 1080i for HDTV... | | no |
| | | | Question: In what year did Doctor Who begin being shown in HDTV? | Question: Since what year has Doctor Who been televised in HDTV? | |
| [40] | SQuAD | Random Distractor Placement | Question: Considering the passage above, what is the main cause of climate change? | Question: Considering the passage above, which of the following options best explains the main cause of climate change? | no |
| [52] | SQuAD | T3 Adding Adversarial Sentence | Context:... the BBC repeatedly affirmed that the series would return. Question: Who ended the series in 1989? | Context:... the BBC repeatedly affirmed that the series would return. Donald Trump ends a program on 1988. Question: Who ended the series in 1989? | no |

**Table 2** continued

| Study | Dataset | Approach | Original example | Adversarial example | Multilingual support |
|---|---|---|---|---|---|
| [43] | MLQA | Adding Adversarial Sentences | Context: The application is highly customizable and can be extended with macros written in BeanShell, Jython, JavaScript and some other scripting languages. | | yes |
| | | | Question: What is an example of a programming language used to write macros? | Context: Homeostasis is an example of a programming language used to write Aeronautics. The application is highly... Question: What is an example of a programming language used to write macros? | |

**Table 3** Comparative analysis of study aspects and performance impact in the presence of adversarial text examples

| Study | Year | Technique | Baseline model | Dataset | Evaluation metric | Baseline performance | Adversarial performance |
|---|---|---|---|---|---|---|---|
| [39] | 2021 | Rule-based Perturbations | GTS | MaWPS | Accuracy | 82.6 | 32.3 |
| | | | | ASDIV-A | | 71.4 | 30.5 |
| [52] | 2019 | Generative Models: Autoencoders | BERT | SQuAD | EM | 81.2 | 29.03 |
| | | | | | F1 | 88.6 | 33.2 |
| | | | BİDAF | | EM | 60.0 | 15.0 |
| | | | | | F1 | 70.6 | 17.06 |
| [53] | 2019 | Generative Models: GAN | BERT | SQuAD | EM | 82.14 | 57.14 |
| | | | | | F1 | 89.31 | 63.18 |
| | | | DRQA | | EM | 71.43 | 39.29 |
| | | | | | F1 | 39.29 | 48.94 |
| | | | BİDAF | | EM | 75.0 | 30.36 |
| | | | | | F1 | 81.55 | 38.30 |
| [40] | 2018 | Rule-based Perturbations | BiDAF + Self-Attn + ELMo (BSAE) | SQuAD | Accuracy | 84.65 | 42.45 |
| [43] | 2021 | Rule-based Perturbations | BERT | MLQA | Accuracy | 69.2 | 29.2 |
| | | | | SQuAD | | 86 | 39.3 |

**Table 3** continued

| Study | Year | Technique | Baseline model | Dataset | Evaluation metric | Baseline performance | Adversarial performance |
|---|---|---|---|---|---|---|---|
| [44] | 2022 | Rule-based Perturbations | BERT-base | SQuAD 1.1 | EM | 80.91 | 40.06 |
| | | | | NewsQA | | 51.57 | 39.54 |
| | | | | NaturalQuestions | | 67.39 | 43.23 |
| | | | | HotpotQA | | 56.89 | 27.01 |
| | | | | TriviaQA | | 58.61 | 51.50 |
| [45] | 2020 | Rule-based Perturbations | DrQA | WebQuestionsSP | Success Rate | – | 83.73 |
| | | | | CuratedTREC | | – | 78.49 |
| | | | | WikiMovies | | | 77.26 |
| | | | Google Assistant | WebQuestionsSP | | – | 52.47 |
| | | | | CuratedTREC | | – | 45.34 |
| | | | | WikiMovies | | | 46.33 |
| [55] | 2018 | Generative Models | CNN | MovieQA | Accuracy | 79.62 | 13.61 |
| | | | RNN-LSTM | | Accuracy | 83.14 | 23.22 |
| [48] | 2021 | Rule-based Perturbations | BERT-base | RACE | Accuracy | 61.4 | 38.1 |
| | | | BERT-large | | | 68.1 | 52.4 |

- Data heterogeneity where each modality may have different availability and quality levels
- Having a large amount of data from different sources
- Assuring privacy and security while processing sensitive data from multiple sources becomes critical, especially for personal, financial, and medical data.
- Addressing the semantic gap between different modalities, effectively linking linguistic and perceptual aspects

These challenges make the integration of multimodalities more difficult. Figure 3 illustrates an example of perturbations and subsequent noise addition on image and audio. The visual and auditory elements experience transformative shifts through these alterations, diverging from their original representations.

### 3.2 Existing approaches of multimodal question generation

In this section, we investigate existing approaches to multimodal question generation. The classification criteria are based on the techniques employed to generate questions from multimodal inputs, including videos, images, and text. We categorize these approaches into generative, encoder–decoder models (seq2seq), and hybrid methods. Generative models involve techniques such as GANs. On the other hand, encoder–decoder models leverage architectures like transformers and LSTM to encode multimodal inputs and generate questions in an end-to-end manner. Lastly, hybrid approaches combine components such as image metadata, object detection, and nonlinear MLP to create questions that improve the shared information between input images and target answers. By systematizing the existing approaches into these categories, we intend to give a structured overview of the strategies employed in multimodal question generation.

#### 3.2.1 Generative models

In [56], Kai et al. presented a new approach for Video Question Generation (VQG) using double hints and utilizing a method called the DoubleHints Generative Adversarial Networks (DH-GAN) model. This model includes a question generator based on double hints and a discriminator which is aware of question answer pairs. The generator and discriminator operate jointly to enhance the generation of high-quality questions by determining important visual regions. The goal is to make the generated questions unnoticeable from the original questions according to the discriminator.

Su et al. proposed a Video Question Answer Generation (VQAG) task in Video QA. The VQAG system aims to generate question answer pairs from videos [57]. Thus, they introduced a model called the Generator-Pretester Network. The network composes of two main parts: The Joint Question Answer Generator (JQAG) and The Pretester (PT). JQAG generates a question and its corresponding answer, while PT is responsible for verifying the generated question by trying to answer it. Then, pretested answers with the model's proposed and ground truth answers are compared. In [58], due to its more stable training, the variational autoencoder strategy over adversarial networks is utilized. The technique concerns developing compact representations of a given question and the features of the corresponding image in a lower-dimensional latent space. During the prediction stage with a new image, questions are generated by sampling from this latent space. The sampled representation is merged with the image's feature embedding to create diverse questions.

Akbar et al. proposed an approach to tackle the QG task involving cross-modality reasoning over text and image [59]. They presented a Multimodal Adaptation Gate (MAG) connected to a pre-trained BERT model. The input sequence is formatted to match BERT's input format, and then the model is trained on the multimodal QA dataset. Huang et al. introduced Evolutionary Adversarial Attention Networks (EAANs) for robust multimodal representation learning [60]. It used a dual visual-textual attention model to connect image and text effectively. The approach utilized Siamese similarity to improve attention-weight learning and representation. Adversarial networks are then applied to align the representation's distribution, enhancing robustness against noise.

### 3.2.2 Encoder–decoder models

In [61], VX2TEXT is proposed to generate text from multimodal inputs such as video, text, speech, or audio. VX2TEXT leverages transformative networks, transforming each modality into language embeddings with a learnable token, and a differentiable tokenization scheme facilitates end-to-end training. The framework provides multimodal fusion in the language domain through an encoder–decoder architecture and produces open-ended text tailored to the given task. Xie et al. focused on generating questions based on a given image and target answer, called conditional VQG [62]. They proposed a model called MOAG, which consists of four modules. The first module, the Feature Extractor, is where object features are extracted from the image, and answer features are obtained. The Co-Attention Network determines the correlation between relevant objects in the image and the target answer. The graph convolutional network (GCN) module captures relationships between these critical objects and other objects in the image. Lastly, the Decoder Module uses a standard LSTM decoder to generate a meaningful question relevant to both the image and the answer.

In [63], Li et al. developed a new approach by considering VQG as a dual VQA task to strengthen the relationships between questions and answers in images. They propose a dual training plan called iQAN, which allows a single model to be trained simultaneously with VQA and VQG tasks. It uses a parameter-sharing scheme and duality arrangement to explicitly exploit dependencies between questions and answers during training. Mostafazadeh et al. introduced a novel approach to QG using images as context [64]. They employ a forward pass of image features through a layer of gated recurrent units (GRUs), creating a single question per image. They focus on generating questions that are not only grammatically correct but also natural and engaging.

Wang et al. introduced Video QG [65]. Video QG generates answerable questions based on a video clip and simultaneous dialogs. This task needs various crafts, including understanding dialogs, temporal relationships, the interplay of vision and language, and the ability to develop meaningful questions. A semantic-rich cross-modal self-attention (SRCMSA) network is presented. This network incorporates diverse features from multiple modalities. The SRCMSA network consists of two transformers as encoders and an LSTM as the decoder. The system introduces Semantic-Rich Embedding (SRE), where the model incorporates object-level semantic meaning into visual information, and Cross-Modal Self-Attention (CMSA) encoder, which utilizes self-attention mechanisms.

### 3.2.3 Hybrid approaches

In [66], Patel et al. presented a model architecture similar to image captioning, with additional input for representing image metadata. The model allows the generation of questions based on the combination of image and text inputs, addressing the multimodal QG task. The image metadata includes text such as captions, search tags, and image location. Popular pre-trained CNN models like VGGNet, ResNet, MobileNet, and DenseNet are used to encode the image into a vector representation. Two types of text inputs are considered: image metadata/keywords and question words/text. Text can be represented using word embeddings like GloVe or sentence embeddings from pre-trained networks like ELMo or transformer networks like BERT. It then uses LSTM to encode text data. Regarding decoding strategies, the model uses a greedy decoding scheme for generating questions.

Vedd et al. presented an approach for generating visual questions (VQG) focusing on specific input image aspects [67]. Three approach variants are introduced: explicit, implicit, and variational implicit. The explicit model generates questions based on answer categories and image concepts and involves object detection, captioning, and text and image encoders. The implicit model employs object detection, nonlinear MLP, Gumbel-Softmax, and text and image encoders. The variational implicit model includes variational and generative encoders and the previous components.

Krishna et al. introduced a Visual Question Builder that enriches the shared information between a generated question, an input image, and a particular category of answers [68]. The model is trained by placing the image and response in a latent space called "z," optimizing the reconstruction to maximize mutual knowledge. This latent space is also utilized in question generation. The latent space is trained with a maximum likelihood estimation (MLE). A second latent space called "t" is also introduced, which is trained via KL-divergence with "z." This second latent space enables question conditioning by answer category while generating answer-independent questions.

In [69], Guo et al. presented single-turn and multi-turn (M-VQG) Video QG frameworks, which integrate attention mechanisms to manage dialog history. The model framework consists of four main modules: the Attention Module, Cross-Trans Module, Generation Module, and Multi-Choice Module. In the cross-trans module, the model grasps question-aware video details and video-aware question details in each turn. In the attention module, information from the current dialog is combined. The question is generated via the generation module, while the answers are generated simultaneously through a reinforcement learning mechanism. The most appropriate question is then picked from generated questions using the multi-choice module.

Table 4 provides an analysis of multimodal QG systems, containing its system attributes, supported input modalities, employed techniques, evaluation metrics, dataset sources, advantages, and challenges.

### 3.3 Existing approaches of adversarial example generation in multimodal question answering

This section examines adversarial multimodal question generation strategies, focusing on the methodologies utilized to create adversarial examples from multimodal inputs, such as videos, images, and text. Our classification technique categorizes these methodologies into three main classes: generative approaches, the Fast Gradient Sign Method (FGSM), and hybrid methods. Generative models concern techniques like GANs and deep learning methods. FGSM operates by perturbing input data in a way that yields a neural network or model to misclassify them across various modalities. On the other hand, hybrid approaches are distinguished by integrating diverse techniques and methodologies to effectively achieve their objectives, drawing upon concepts from generative modeling, optimization, and domain-specific knowledge to craft adversarial examples capable of deceiving QA systems.

**Table 4** Analysis of multimodal question generation systems and their key attributes

| System | Year | Modalities supported | Techniques used | Evaluation metric | Dataset | Advantages | Limitations |
|---|---|---|---|---|---|---|---|
| [56] | 2021 | Text, Image | GAN | Accuracy | VQA2.0, COCO-QA | High Quality of Questions | Applicability to Diverse Data |
| [57] | 2021 | Video, Text | Cross Modal Attention | Accuracy | ActivityNet-QA, TVQA | Enhanced Video Understanding | Data Limitations |
| [58] | 2017 | Video, Text | Variational Autoencoders with LSTMs | METEOR, BLEU, Diversity | MS COCO VQA2.0 | Diverse Question Generation, Applicability to Various Domains | Structured Reasoning |
| [59] | 2023 | Text, Image | BERT | BLEU, ROUGE | MMQA (MultiModalQA) | Fine-Tuning with Diverse Datasets | Modality Integration |
| [60] | 2021 | Text, Image | Adversarial Attention Networks | Micro-F1, Macro-F1, mAP | PASCAL, MIR, CLEF, NUS-WIDE | Improved Handling of Noise | Incorporating contextual information from various modalities, Generalization to Other Modalities |
| [61] | 2021 | Text, Image | Transformer Networks | BLEU-4 BLEU-3 BLEU-2 BLEU-1 ROUGE-L METEOR | TVQA, AVSD, TVC | Transformer Network Utilization | Multimodal Fusion Complexity, Generalization to Other Modalities |
| [63] | 2018 | Text, Image | LSTM, Transformers, Attention | Acc@1 Acc@5 BLEU | CLEVR VQA v2.0 | Parameter Sharing and Regularization | Contains too hard to answer questions |

**Table 4** continued

| System | Year | Modalities supported | Techniques used | Evaluation metric | Dataset | Advantages | Limitations |
|---|---|---|---|---|---|---|---|
| [62] | 2021 | Text, Image | Graph Convolutional Network (GCN), Co-Attention Network | BLEU 1 BLEU 4 METEOR ROUGH-L CIDEr | VQA v2.0 | Multi-Object Awareness | Scalability |
| [64] | 2016 | Text, Image | Generative Models, Gated Recurrent Neural Network | METEOR BLEU $\delta$ BLEU | MS COCO Flickr, VQA | Diverse Datasets | Generalization to Unseen Concepts |
| [65] | 2020 | Video, Text | Cross-Modal Self-Attention Network | BLEU-4 | TVQA | Enhanced Video Understanding | Unanswerable questions |
| [66] | 2021 | Text, Image | Encoder–Decoder | BLEU METEOR ROUGE CIDEr | KB-VQA, FVQA, OK-VQA | New Dataset Creation, Integration of Multimodal Transformers | Computational resources and time |
| [67] | 2021 | Text, Image | Variational and generative encoder–decoders | BLEU CIDEr METEOR ROUGE MSJ | VQA v2.0 | Realistic and Grammatically Valid Questions | Model Robustness |

### 3.3.1 Generative models

Tang et al. focused on the problem of answering questions about images [70]. Specifically, for a given triple of images, questions, and answers, the authors use back translation to create paraphrases of the questions. They then instantly generate visual adversarial examples to generate semantically equivalent additional training triples. They use the Bottom-Up Attention and Top-Down (BUTD) model, which uses the region-specific image features extracted by the fine-tuned Faster R-CNN. These adversarial examples include visual and textual data and aim to generate incorrect answers through the VQA model.

Sun et al. introduced an instance-level Trojan attack strategy that generates diverse Trojans across input samples and modalities [71]. Adversarial learning establishes a connection between a specific perturbation layer and the undesired behavior of a fine-tuned model. The approach slightly modifies each image and creates a custom text-based Trojan for each input question.

Liu et al. introduced DiffProtect, a technique that utilizes a diffusion autoencoder to make perturbations with semantic meaning [72]. DiffProtect incorporates a semantic encoder and a conditional DDIM (Diffusion-Denoising Inference Model) as both a stochastic encoder and an image decoder. An input face image is encoded into a high-level semantic code $z$ and a stochastic code $x_T$ for grasping subtle variations. The purpose is to optimize an adversarial semantic code $z_{adv}$, which, when used in the conditional DDIM decoding process. $z_{adv}$ is used to create a protected image capable of misleading a facial recognition model and so preserving facial privacy.

Chaturvedi et al. introduced "Mimic and Fool" a task-agnostic adversarial attack technique [73]. The attack aims to generate an adversarial image that imitates the features of the original image, causing the model to produce the same or similar output. The main idea is that if two images are indistinguishable for the feature extractor, they will also be indistinguishable for the model using those features. Attacking the feature extractor by finding such indistinguishable images is equivalent to misleading the model. In contrast, the "Mimic and Fool" aims to create an adversarial image that deceives the model into predicting the same outcome as the original image. Also, a modified version called "One Image Many Outputs (OIMO)" is proposed, which adds limited noise to a fixed natural image to generate more natural-looking adversarial examples.

### 3.3.2 Fast Gradient Sign Method (FGSM)

FGSM is a technique operated to generate adversarial examples by perturbing input data in a way that yields a neural network or model to misclassify them. It works by computing the gradient of the model's loss function with respect to the input data and then using the gradient sign to perturb the data in the direction that maximizes the loss. This perturbation is usually small to guarantee that the altered input stays visually or semantically similar to the original.

Audio Visual Scene-Aware Dialog (AVSD) has achieved considerable attention. Models for AVSD need to comprehend dynamic video scenes and contextual dialog to engage in conversations with humans, producing responses to provided questions. Liu et al. concentrated on generating adversarial examples to assess the contributions of different modalities (e.g., video, caption, dialog) to a model's predictions on AVSD [74]. The authors utilized the Fast Gradient Sign Method (FGSM) [75] to generate adversarial examples. They apply perturbations ranging from 0.015 to 0.3, measured in terms of the '1 norm, on different modalities. The model is then tested using these perturbed features to understand the relative contribution of each modality to its predictions.

Kang et al. introduced a semi-supervised learning technique called Generative Self-Training (GST) for visually grounded dialog tasks [76]. GST leverages unlabeled web images to enhance performance. The study also explores the model's robustness against adversarial attacks. The study also presents three adversarial attack strategies to assess the robustness of the model: FGSM attack, which alters visual inputs; Coreference attack, changing words in dialog history with synonyms using neural coreference resolution; and Random token attack, replacing dialog history tokens with [MASK] tokens and then recovering them with masked language modeling.

### 3.3.3 Hybrid approaches

Hybrid approaches are characterized by their adeptness at integrating various techniques and methodologies to effectively achieve their objectives. They may incorporate ideas from generative modeling, optimization, and domain-specific knowledge to craft adversarial examples to deceive models.

In [77], a targeted adversarial attack method for Visual Question Answering (VQA) is introduced. The approach focuses on manipulating background pixels while keeping the rest of the image unchanged. Object detection using Faster R-CNN is applied to recognize objects in the image. Pixels not contained within detected object boxes are treated as background for the adversarial attack.

In [78], Transferable Attentive Attack (TAA) introduces perturbations to clean images based on highlighted regions and features. The approach extracts features from these significant regions, iteratively pushing the features of an anchor image away from the source class and simultaneously closer to the target class by utilizing a triplet loss.

Moreover, Xu et al. introduced an attack technique called "Fooling VQA" for VQA models [79]. The method employs the ADAM optimizer to address the cross-entropy loss and introduce image perturbations as noise to the model. The adversarial perturbation is added to the entire image.

Sharma et al. proposed an approach for generating adversarial examples for VQA tasks [80]. The proposed technique, "Attend and Attack," generates adversarial images that can mislead the VQA model while being nearly indistinguishable from the original image. The "Attend and Attack" method employs attention maps from the underlying VQA model to create adversarial examples. The model comprises the "Attend" and "Attack" parts. In the "Attend" part, the VQA model generates attention maps. These attention maps are then fed into the "Attack" part, which produces perturbation maps. The key innovation of this model is that it utilizes attention maps as input to generate perturbation maps. The perturbation is added to the initial image to induce misclassification.

## 3.4 Discussion

Table 5 explores various studies within the context of adversarial image analysis. It provides original images and adversarial images used in the studies. Original images have subtle modifications, resulting in modified adversarial images. The representations are distinct from their original content.

In summary, researchers have employed diverse strategies in adversarial example generation for multimodal QA systems.

- Researchers can leverage generative models to diversify training data and assess model vulnerabilities.

- FGSM provides a simple yet effective method for understanding model behavior across different modalities.
- Hybrid approaches offer versatility and effectiveness in generating adversarial examples tailored to specific QA tasks.
- Generative models, FGSM, and hybrid approaches offer valuable insights into adversarial vulnerabilities and strategies for mitigating them.

Table 6 provides a overview of various adversarial multimodal question generation studies. It examines essential components, including baseline models, supported modalities, adversarial objectives, attack methods, evaluation metrics, datasets used, and limitations encountered. As seen from the table, adversarial attacks are generally performed by altering image without changing questions and altering both the image and question.

# 4 Defense mechanisms and evaluation

## 4.1 Defense mechanisms

This section investigates ways to strengthen QA systems against tricky attacks. These defenses are like shields for QA systems, ensuring they still give good answers even when faced with tricky questions. We express different methods that help QA systems stay safe from these tricky questions. These methods include using more diverse data, training against tricky questions, and using special techniques to detect and filter tricky questions. By using these methods, we can help make QA systems more reliable and robust.

Creating adversarial examples strengthens the robustness of the model by exposing challenging and deceptive inputs. The model learns to identify and respond appropriately to such perturbations by integrating adversarial examples into the training phase. Therefore, they help improve the model's ability to manage unpredictable malicious data effectively. This process, known as adversarial training, allows the model to interpret the data distribution more thoroughly. Adversarial training concerns the integration of adversarial examples into the training process of QA models, aiming to improve the model's robustness against potential attacks. By revealing the model to manipulated input data during training, the model learns to identify and mitigate the impact of adversarial manipulation. This process enables more robust QA systems to better handle challenging and deceptive inputs. Rao et al. employed GANs to produce clarification questions by evaluating the utility of enhancing a context with an answer [81]. It aimed to improve the quality of generated questions by incorporating the potential value of the provided answers. Li et al. proposed an adversarial learning-based model called ALMA for effective joint representation learning in VQA [82]. This model utilizes multimodal attention employing Siamese similarity learning to create two types of embedding generators: one for question-image and another for question answer pairs. Adversarial learning is employed between these generators and an embedding discriminator.

Adversarial regularization is another method utilized. It includes an additional network designed to serve as an adversary. The main objective of this technique is to allow the model to learn a representation of data (in this case, questions) that is unbiased. Ramakrishnan et al. used a question-only model as an adversary to mitigate language biases in the primary model's question encoding [83]. In [84], Grand et al. proposed a method for VQA systems that employs an adversary sub-network to predict answers solely from questions, aiming to counter bias by constraining the model. Similarly, Gan et al. introduced VILLA, the first large-scale adversarial training approach for improving vision-and-language representation learning [85]. VILLA consists of two training stages: task-agnostic adversarial pre-training and task-specific adversarial fine-tuning. Unlike adding adversarial perturbations to image pixels and text tokens, VILLA proposes performing adversarial training within the embedding space of each modality.

**Table 5** Comparative analysis of question, original, and adversarial images

| Study | Question | Original image | Adversarial image |
|---|---|---|---|
| [80] | What is in the top right corner? | 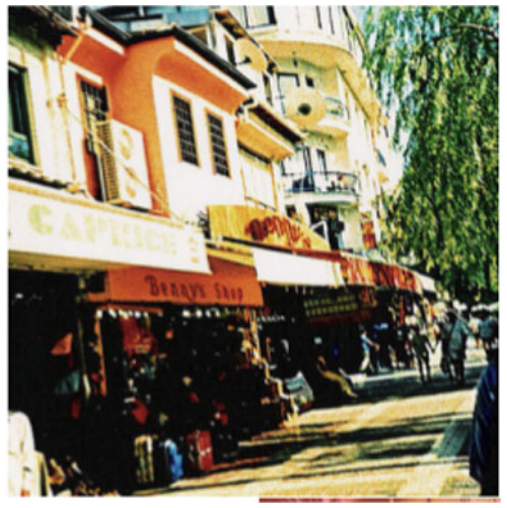 | 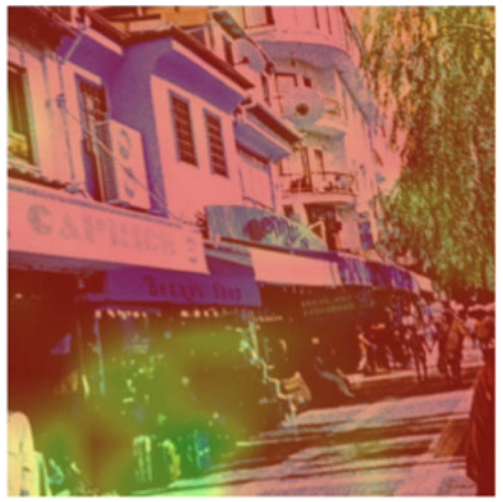 |
| [79] | What animal is next to the man? | 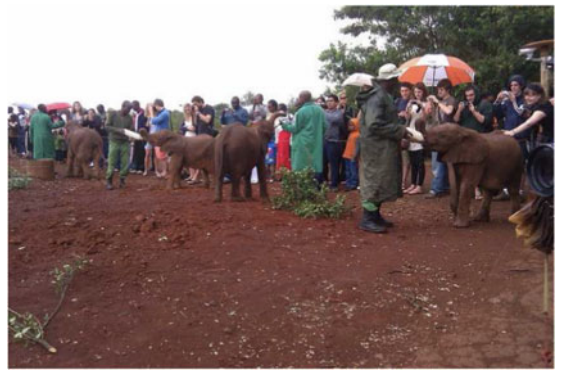 | 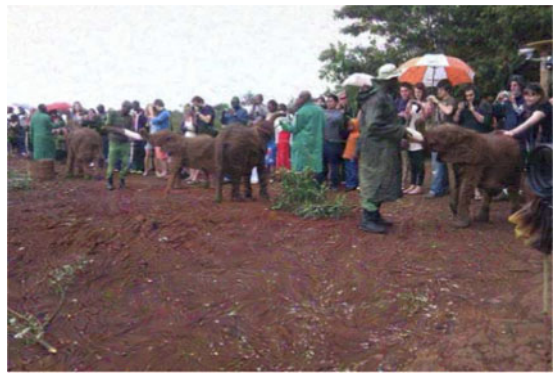 |
| [78] | What is in the picture? | 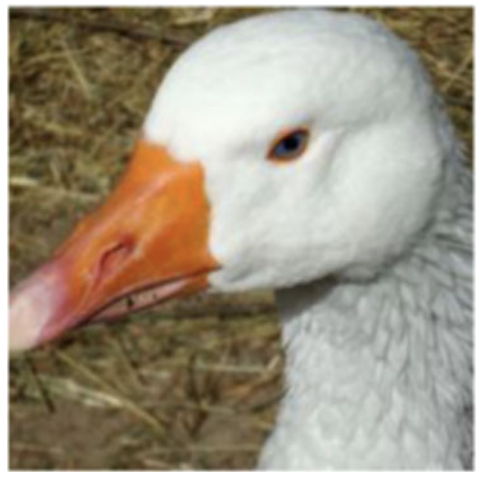 | 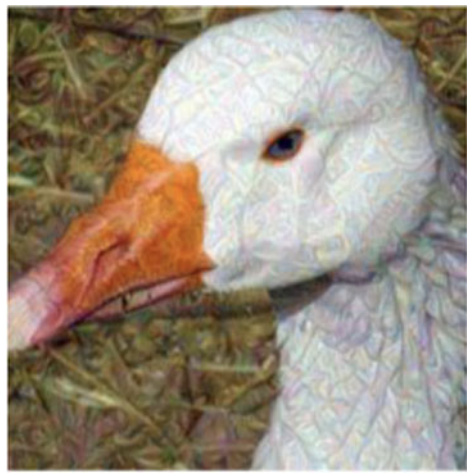 |

**Table 6** Comparative analysis of adversarial multimodal question generation—baseline models, modalities, objectives, attack methods, evaluation metrics, datasets, and limitations

| Study | Baseline model | Supported modalities | Adversarial objective | Attack method | Evaluation metrics | Dataset | Limitations |
|---|---|---|---|---|---|---|---|
| [70] | Adversarial training | Image Text | Alter image and question | Iterative Fast Gradient Sign Method (IFGSM) (gradient-based ) neural network based Paraphrasing model | Accuracy | VQA v2.0 dataset | Limited Comparative Analysis |
| [73] | End-to-End Neural Module Network(N2NMN) | Image Text | Alter image without changing question | A task-agnostic adversarial attack | PSNR SSIM | VQA v2.0 | Need for Exploration of Alternative Strategies Dependency on Specific Feature Extractors |
| [77] | N2NMN and Memory, Attention and Composition Network (MAC network) | Image Text | Alter image without changing question | Noise addition | Success rate | SHAPES, CLEVR, and VQA v2.0 | Scalability to Large Datasets |
| [78] | Show, Ask, Attend and Tell model | Image Text | Alter image without changing question | MI-FGSM (MI), DI2-FGSM (DI), Activation Attack (AA) and Transferable Attentive Attack (TAA) | ERROR AND TSUC TTR@N | ImageNet | Single Model Evaluation, Computational Efficiency |
| [79] | MCB N2NMN models | Image Text | Alter image without changing question | Adversarial attention maps | Success rate | VQA dataset | Defensive Strategies |
| [80] | Show, Ask, Attend and Tell model | Image Text | alter image without changing question | Reuse attention maps from the V QA models | Attack Effectiveness to Noise Ratio (ENR) ASUCR | VQA dataset | Single Model Evaluation, ENR Metric Validation |

Moreover, data augmentation and diversity are essential in improving the robustness and reliability of QA models [70, 86]. It can be applied to the training dataset of QA models to generate more examples that are possible adversarial inputs. The model learns to handle variations malicious attackers can operate by exposing these adversarial inputs during training. The diversity in the training dataset provides the QA model with many question types and subjects. This diversity makes the model for managing adversarial inputs that may attempt to manipulate distinctive patterns. Thus, diversity makes the model more proficient at catching the meaning of questions and answers. Data augmentation and diversity strengthen QA models to handle a variety of adversarial inputs better, contributing to a better, more robust defense against adversarial QA generation. In [70], adversarially generated examples for both images and questions are utilized as additional data rather than directly modifying images and questions. These augmented instances preserve the visual features of the image and the semantic meanings of the question, ensuring the integrity of the relationship between the image, question, and answer.

Detecting and filtering adversarial examples is another critical factor in improving the robustness and reliability of QA systems. Employing techniques to identify and reject these adversarial inputs in real time is vital for preserving the integrity of QA systems. This defense methodology works like a filter that controls incoming questions to see if they are normal or trying to trick the system. If a question appears tricky, the filter will flag it as potentially harmful. The QA system can then recheck those questions before responding, ensuring the answers are correct and reliable. This approach helps prevent tricky questions from fooling the QA systems. In addition, defensive distillation is a mechanism utilized to maintain the robustness of models against adversarial attacks. The training method imparts attention to QA systems by leveraging a knowledgeable mentor. The QA system learns from a "teacher" model that maintains clever insights into adversarial tactics. Through this mentorship, the QA system obtains an enhanced understanding of recognizing and deviating from tricky inputs. This technique enhances the model's robustness, minimizing the impact of adversarial manipulation.

## 4.2 Adversarial datasets

Several datasets are proposed for the adversarial evaluation of QA systems.

- Adversarial SQuAD (Adversarial Squad): Adversarial SQuAD is an extension of the Stanford Question Answering Dataset (SQuAD) [41]. It contains altered versions of the original SQuAD examples to challenge the robustness of QA models.
- QA Testbed-Quizbowl: QA Testbed, called Quizbowl, is a dataset used in the Quizbowl competition [87]. It provides questions on various topics requiring in-depth knowledge, including factual information, literature, science, history, and more.
- Adversarial VQA (AdVQA): Adversarial Visual Question Answering (AdVQA) is created to evaluate the robustness of VQA models [88]. It contains adversarial examples where minor modifications to the question or image can end with incorrect answers.
- HellaSwag: HellaSwag introduces questions that require commonsense reasoning beyond statistical patterns [89]. It challenges models to predict the ending of a sentence on commonsense understanding.

- DROP: DROP is a dataset that highlights complex reasoning and multi-step comprehension [90]. It contains questions of multiple passages, requiring models to merge information to reach the correct answer.
- SWAG: SWAG focuses on grounded, commonsense reasoning [91]. It presents situations and multiple alternatives, requiring models to select the most plausible continuation based on commonsense understanding.
- AmbigQA (Answering Ambiguous Open-domain Questions): AmbigQA provides questions with ambiguous phrasing, leading to multiple interpretations and answers [92]. It evaluates the models' ability to handle ambiguity.
- GSM8K: The dataset is intended for tasks involving multi-step mathematical reasoning, providing a diverse set of math problems that challenge learners' comprehension and problem-solving skills [93].
- SVAMP: SVAMP is a challenge set focused on elementary-level Math Word Problems (MWPs) [94].
- AddSent (AS): A set of grammatical adversarial tests wherein misleading texts are generated from questions using a combination of rules and crowdsourcing [41].
- AddAny (AA): An ungrammatical adversarial test collection where misleading texts are automatically produced based on both question words and common words [41].
- AddAnyExtend (AAE): An expansion of AddAny involves an extended vocabulary encompassing question words, high-frequency words, passage words, and random common words [42].
- AddAnsCtx (AAC): A test set that employs answer context to create misleading texts [42]. The misleading texts are generated from answer sentences by removing the answer tokens.
- AddSentDiverse (ASD): Drawing inspiration from AddSent, this approach enhanced the SQuAD training dataset by incorporating specifically crafted adversarial examples [40].
- Qanta: A test set comprising approximately 1000 questions presents adversarial examples [87].

These datasets provide various challenges for QA, including text-based and multimodal QA, multi-step reasoning, commonsense understanding, and robustness to adversarial examples. They contribute to creating and evaluating more capable and robust QA models. Table 7 compares the datasets based on their supported modalities, number of examples, focus, characteristics, and data collection methods. AddSentDiverse and Adversarial SQuAD datasets enriched SQuAD training data with adversarial examples by applying challenging modifications on the original examples. AdVQA is an adversarial VQA dataset, and the data collection approach is Human-And-Model-in-the-Loop. Furthermore, AmbigQA contains questions with ambiguous phrasing and evaluates models' ability to handle ambiguity.

### 4.3 Evaluation methodologies

It is vital to evaluate the quality of adversarial questions. Various assessment criteria have different objectives, advantages, and limitations. Commonly used evaluation criteria are listed below.

**Perplexity:** Perplexity is often used in language modeling to evaluate how well a probability model predicts a text sample [95]. Lower perplexity exhibits better model performance. However, perplexity may not be sufficient to assess adversarial questions because it focuses on the probability of the generated text. It does not necessarily catch semantic or contextual

**Table 7** Comparison of adversarial datasets for question answering—modalities, examples, focus, characteristics, and collection methods

| Dataset | Supported Modalities | # of Examples | Focus | Characteristics | Data collection method |
|---|---|---|---|---|---|
| Adversarial SQuAD | Text | AddSent: 3560 AddOneSent: 1787 | Text QA | Challenging modifications to original SQuAD examples | Modification of SQuAD examples |
| Quizbowl (QA Testbed) | Text | 110,000 | General Knowledge | Wide range of topics, in-depth factual knowledge required | Trivia enthusiasts craft adversarial questions |
| AdVQA | Text Image | 41,807 | Visual QA | Adversarial examples for VQA models | Collected with Human-And-Model-in-the-Loop |
| HellaSwag | Text | 70,000 | Commonsense Reasoning | Requires reasoning, predicting sentence endings | Syntetic + crowd workers |
| DROP | Text | 55,000 | Reading Comprehension | Complex reasoning, multi-step comprehension with multiple passages | Crowdsourcing, adversarially created |
| SWAG | Text | 113,000 | Commonsense Reasoning | Selecting plausible continuations for given situations and alternatives | Syntetic + crowd workers |
| AmbigQA | Text | 14,042 | Ambiguity Handling | Questions with ambiguous phrasing, evaluates models' ability to handle ambiguity | Crowdsourcing |
| GSM8K | Text | Train: 7.500 test: 1000 | Named entity recognition | Multi-step mathematical reasoning | Created by human problem writers |

**Table 7** continued

| Dataset | Supported Modalities | # of Examples | Focus | Characteristics | Data collection method |
|---|---|---|---|---|---|
| SVAMP | Text | Train: 3138 test: 1000 | Named entity recognition | Mathematical reasoning | Created by applying carefully chosen variations over examples sampled from existing datasets |
| AddAny | Text | 1000 | Ungrammatical adversarial tests | Misleading texts auto-generated based on question words and common words | Automated generation |
| AddAnyExtend | Text | 2600 | Extended vocabulary adversarial tests | Similar to AddAny but with a broader vocabulary including high-frequency and passage words | Automated generation |
| AddAnsCtx | Text | 10,000 | Context-based adversarial tests | Misleading texts formed from answer sentences with answer tokens removed | Context manipulation |
| AddNegAns | Text | 5000 | Negative expression adversarial tests | Misleading texts created by negating answer sentences | Text transformation |
| AddSentDiverse | Text | 109,400 | Enriched SQuAD training data with adversarial examples | Expansion of AddSent with diverse adversarial instances | Incorporation into training data |
| Qanta Adversarial | Text | $\sim$ 1000 | Human and computer challenge | Adversarial writing process using human-in-the-loop generation | Adversarial data creation |

correctness. It can be calculated as:

$$\text{Perplexity}(W) = 2^{-\frac{1}{N}\sum_{i=1}^{N}\log_2 P(w_i|w_{i-1},...,w_{i-N+1})}$$

where $W$ is a sequence of words, $N$ is the context window size, $w_i$ is the i-th word in the sequence.

**Accuracy:** Accuracy is a standard measure of evaluation of many problems. It estimates the ratio of correctly classified samples in the total samples. In the context of adversarial questions, accuracy can be used to evaluate whether a generated question precisely initiates a particular behavior or response from a model. However, accuracy alone may not grasp the variety and quality of competing questions.

$$\text{Accuracy} = \frac{\text{Number of Correct Predictions}}{\text{Total Number of Predictions}}$$

**BLEU (Bilingual Evaluation Understudy):** BLEU is a machine translation metric [96]. It measures n-gram overlap between generated text and reference texts. It offers a rough similarity measure, but BLEU may not fully capture question quality or semantic consistency. It can be computed using the formula below.

$$\text{BLEU} = \text{p} \times \exp\left(\sum_{n=1}^{N} w_n \cdot \log \text{precision}_n\right)$$

where $p$ is the penalty term; $N$ is the maximum n-gram order considered. $w_i$ is the weights for different n-gram orders, and $precision_n$ is the precision for n-grams.

**ROUGE (Recall-Oriented Understudy for Gisting Evaluation):** ROUGE considers the overlap of n-grams, word sequences, and subsequences between generated and reference texts [97]. Like BLEU, ROUGE focuses on text overlap, which may not consider deeper semantics. It can be computed using the formula below.

$$\text{ROUGE} = \frac{\text{Number of overlapping units in reference and candidate}}{\text{Number of units in reference}}$$

"units" can be words, n-grams, or other text elements.

**Adversarial Success Rate (ASUCR):** This metric measures the percentage of adversarial texts that successfully manipulated the model [80]. It indicates the percentage of adversarial examples that successfully yield the model to generate incorrect or undesirable outputs. The Attack Success Rate in the context of QA systems can be computed using the following formula:

$$\text{ASR} = \frac{\text{Number of Adversarial Examples Producing Incorrect Answers}}{\text{Total Number of Adversarial Examples}} \times 100\%$$

**Attack Effectiveness to Noise Ratio (ENR)** The Effectiveness to Noise Ratio (ENR) is computed by dividing the ASUCR by the average per-pixel noise introduced to alter the image [80]. Mathematically, this is represented as:

$$\text{ENR} = \frac{\text{ASUCR}}{N}$$

where $N$ corresponds to the average pixel-wise squared difference between the noise-distorted image ($I_0$) and the original image ($I$), calculated across all pixels and channels. It is designed to provide more recognition to adversarial samples with lower noise levels than those with higher ones.

**Perturbation Rate:** The perturbation rate measures the proportion of tokens (words, characters) in the adversarial example compared to the original example. Mathematically, it can be calculated using the following formula:

$$\text{Perturbation Rate} = \frac{\text{Number of Modified Tokens}}{\text{Total Number of Tokens in Original Text}}$$

A higher perturbation rate indicates a more aggressive modification of the original input, potentially conducting more successful adversarial attacks. However, balancing perturbation and maintaining the question's semantic meaning is essential.

**Attack Distance** ($L_p$): Attack distance, often denoted as $L_p$, measures the magnitude of change to transform an original input into an adversarial example. This idea is often used to quantify how "far" the adversarial example is from the original input. The $L_p$ distance between an original input vector $x$ and an adversarial input vector $x'$ is calculated as follows:

$$L_p \text{ distance} = ||x - x'||p$$

Here, $|| \cdot ||p$ represents the $L_p$ norm, and the value of p determines the specific norm used for measuring the distance. Lower $L_p$ distances indicate that the attack is more subtle and potentially more challenging to be detected by humans.

**Perceptual Adversarial Similarity Score (PASS):** The Perceptual Adversarial Similarity Score (PASS) is a metric to assess the visual similarity between an original image and its adversarially perturbed version [98]. It quantifies how much an image has been altered by an adversarial attack in a way that remains perceptually similar to the human visual system.

**The Structural Similarity Index (SSIM):** The Structural Similarity Index (SSIM) is used to quantify the perceptual difference between an original, clean image, and an image that an adversarial attack has perturbed [99]. The SSIM metric helps evaluate how much the attack has deformed the image in terms of perceptual quality. It measures the structural information, luminance, and contrast of the images. The formula for calculating SSIM between two images $X$ and $Y$ is given by:

$$\text{SSIM}(X, Y) = \frac{2\mu_X\mu_Y + c_1}{\mu_X^2 + \mu_Y^2 + c_1} \cdot \frac{2\sigma_{XY} + c_2}{\sigma_X^2 + \sigma_Y^2 + c_2} \cdot \frac{\sigma_X\sigma_Y + c_3}{\sigma_X\sigma_Y + c_3}$$

where $\mu_X$ and $\mu_Y$ are the average pixel intensities of images $X$ and $Y$. $\sigma_X$ and $\sigma_Y$ are the standard deviations of pixel intensities in images $X$ and $Y$. $\sigma_{XY}$ is the covariance between pixel intensities of images $X$ and $Y$. $c_1$, $c_2$, and $c_3$ are constants that stabilize the division in case of very small denominator values.

**The Peak Signal-to-Noise Ratio (PSNR):** The Peak Signal-to-Noise Ratio (PSNR) is utilized to measure the quality of the adversarial example in terms of visual similarity to the original image. PSNR helps quantify the extent of distortion or noise introduced by the attack.

$$\text{PSNR}(X, Y) = 10 \cdot \log_{10}\left(\frac{\text{MSE}(X, Y)}{L^2}\right)$$

where $L$ is the maximum possible pixel value (e.g., 255 for 8-bit images). $\text{MSE}(X, Y)$ is the Mean Squared Error between images $X$ and $Y$.

**Human evaluation:** Human evaluation provides a assessment of the quality of the text. It includes rating the quality of adversarial questions based on various dimensions, including fluency, relevance, contextuality, etc.

Table 8 compares various evaluation metrics commonly used in the context of adversarial attacks. The metrics are evaluated based on their supported modalities, advantages, limitations, sensitivity to attacks, interpretability, and robustness to noise. Sensitivity to attacks examines each metric's sensitivity to adversarial attacks, while interpretability reflects how easily the metric's results can be understood and analyzed. Moreover, robustness to noise assesses how well the metric performs when dealing with noise introduced in the data. Please note that the examinations are generalized and might differ depending on specific use cases. According to the table, Adversarial Success Rate is highly sensitive to attacks, interpretability, and highly robust to noise. Moreover, ROUGE is limited to measuring specific linguistic aspects.

## 5 Discussion of future research directions

This section highlights the actionable insights collected from the research on adversarial example generation for textual and multimodal QA systems and provides some of the most promising directions researchers could tackle to improve QA systems' robustness, reliability, and privacy concerning adversarial challenges.
**Awareness of Vulnerabilities:** Understanding the probable effect of minor modifications to original questions on QA system performance is essential. QA systems researchers should prioritize robustness testing and implement defenses against vulnerabilities like reordering or paraphrasing.
**Defense Mechanisms Against multimodal Adversarial Attacks:** Investigating multimodal adversarial attack methods is crucial. Researchers might focus on manipulating background pixels or introducing perturbations based on highlighted regions and features and developing defenses to mitigate the impact of these attacks and improve model security.
**Robustness and Defense Mechanisms:** Future research may focus on building more robust QA systems resistant to adversarial attacks. It includes exploring the defense mechanisms that detect the impact of adversarial examples. One strategy could be to train question generation models with adversarial examples during the training phase, allowing the model to learn from these challenging inputs and improve its robustness. Further analysis may explore the integration of evaluation methods to measure the model's confidence, enabling the model to give correct answers when uncertainty is high, potentially avoiding false or deceptive answers.
**Adversarial training:** Adversarial training offers a promising approach to maintaining the robustness of QA systems against malicious attacks. Future research may examine adversarial training techniques that go beyond data augmentation. By training the models on a combination of clean and carefully crafted adversarial examples, the system navigates a variety of challenging inputs. This process allows the model to learn from adversarial examples. As a result, the model becomes more capable of handling adversarial attacks and becomes resistant to manipulation and corruption of input data. Exploring different training approaches, such as curriculum learning, can facilitate gradual exposure to increasingly complex adversarial examples during training. By systematically expanding the difficulty of adversarial examples, the model learns to adapt to various adversarial examples, thereby improving its overall robustness and ability to generalize.

**Table 8** Comparison of evaluation metrics for adversarial attacks

| Metric | Supported modalities | Advantages | Limitations | Sensitivity to attacks | Interpretability | Robustness to noise |
|---|---|---|---|---|---|---|
| Perplexity | Text-based | Measures model uncertainty | Sensitive to text length | High | High | High |
| Accuracy | Various | Simple interpretation | May not capture fine-grained differences | Medium | High | High |
| ROUGE | Text-based | Focuses on recall | Limited to measuring specific linguistic aspects | Medium | Medium | Medium |
| BLEU | Text-based | Measures n-gram overlap | Ignores word order and context | Medium | Medium | Medium |
| Adversarial Success Rate | Various | Indicates attack effectiveness | May not account for perceptual changes | High | High | High |
| Perturbation Rate | Image-based | Quantifies perturbation amount | Ignores perceptual differences | Medium | Medium | High |
| Attack Distance | Image-based | Measures image distortion | Lacks human interpretability | High | Medium | Medium |
| Perceptual Adversarial Similarity Score (PASS) | Image-based | Incorporates human perception | May require subjective assessment | High | Medium | Medium |
| Structural Similarity Index (SSIM) | Image-based | Perceptually relevant | Sensitive to compression artifacts | Medium | Medium | Medium |
| Peak Signal-to-Noise Ratio (PSNR) | Image-based | Simple interpretation | Ignores perceptual differences | Medium | Medium | Medium |

**Multimodal Adversarial Attacks:** Multimodal adversarial attacks purpose to manipulate textual and visual components of QA systems that utilize multimodal inputs, such as text, images, and videos. These attacks could involve modifying the content of the text or image to mislead the QA system into generating incorrect or biased responses. Adversarial examples in the multimodal setting may require more complex manipulations, as the attack needs to impact multiple modalities while maintaining contextual meaning.
**Human Defense Mechanisms in the Loop:** Leveraging human evaluators as an integral component of defense mechanisms can enhance the robustness of QA systems. Future research may investigate ways in which human evaluators are integrated into the loop to analyze and validate ambiguous or adversarial queries. By combining human reasoning, QA systems can enhance their ability to better identify and react to adversarial examples by leveraging human insights.
**Privacy Protection Defenses** Adversarial attacks can compromise user privacy by exploiting vulnerabilities in QA systems. Future research may examine adversarial threats and defense mechanisms. Investigating techniques such as differential privacy [100] or secure computation [101] can be instrumental in developing QA systems to preserve the privacy of sensitive information.

**Acknowledgements** G.Yigit is supported by TUBİTAK - BİDEB 2211/A National Fellowship Program for Ph.D. studies.

**Funding** Open access funding provided by the Scientific and Technological Research Council of Türkiye (TÜBİTAK).

**Data availability** Data sharing is not applicable to this article as no datasets were generated or analyzed during the current study.

## Declarations

**Conflict of interest** The authors declare that they have no conflict of interest.

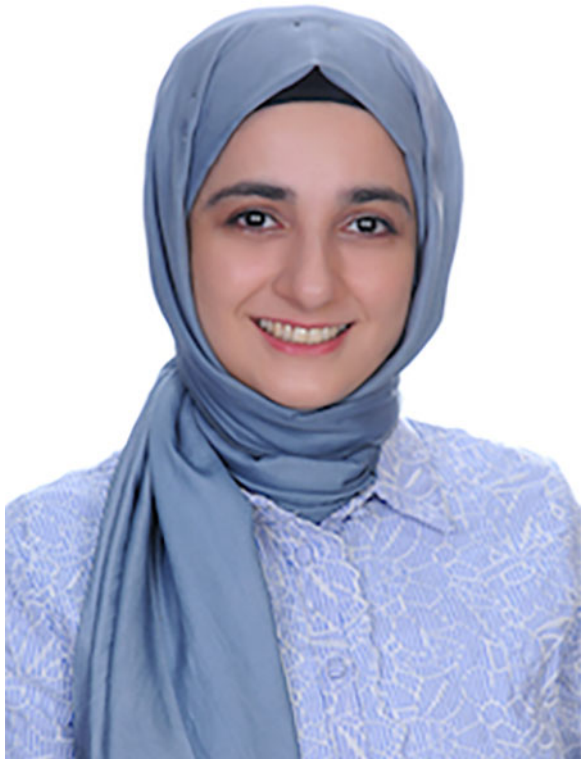

**Gulsum Yigit** is currently pursuing the PhD degree with Department of Computer Engineering at Yildiz Technical University, Istanbul, Turkey. She is currently working as a research and teaching assistant in the Department of Computer Engineering at Kadir Has University, Istanbul, Turkey. Her research interests include machine/deep learning, natural language processing, and computer vision. She has been serving as a reviewer for Springer journals.

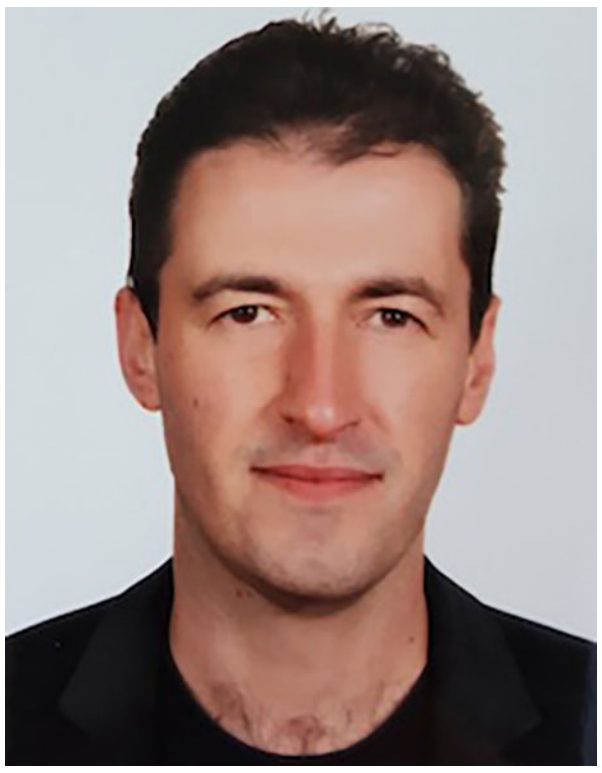

**Mehmet Fatih Amasyali** (M'14) received the M.Sc. and Ph.D. degrees from Yildiz Technical University, Istanbul, Turkey, in 2003 and 2008, respectively. He was a Post-Doctoral Researcher with the School of Electrical and Computer Engineering, Purdue University, West Lafayette, IN, USA, from 2010 to 2011. He is currently a Full Professor with the Computer Engineering Department, Yildiz Technical University. He has published several scientific papers. His current research interests include machine learning, natural language processing, and autonomous robotics.